\pdfoutput=1

\documentclass[11pt,dvipsnames]{article}

\usepackage[preprint]{acl}
\usepackage{multirow,makecell}
\usepackage{arydshln}
\usepackage{graphicx} 
\usepackage{tabularx}
\usepackage{booktabs}
\usepackage{times}
\usepackage{latexsym}

\usepackage{xcolor}
\usepackage{soul}  

\usepackage[T1]{fontenc}

\usepackage[utf8]{inputenc}

\usepackage{microtype}

\usepackage{inconsolata}
\usepackage{kotex}

\usepackage{graphicx}
\usepackage{amsmath}
\usepackage{amsthm,amssymb,multirow}
\usepackage{algpseudocode}
\usepackage{algorithm}
\usepackage{enumitem}
\setlist[description]{%
  leftmargin=10pt,
  labelindent=10pt,
  itemsep=0pt,    
  font=\normalfont\bfseries
}

\colorlet{lightblue}{Cerulean!40}

\newcolumntype{L}{!{\vrule width 1pt}}  
\newcolumntype{R}{!{\vrule width 1pt}}  

\title{AmpleHate: Amplifying the Attention for Versatile Implicit Hate Detection}

\author{
 \textbf{Yejin Lee},
 \textbf{Joonghyuk Hahn},
 \textbf{Hyeseon Ahn} \and
 \textbf{Yo-Sub Han}\thanks{Corresponding author.}
 \\
  Yonsei University, Seoul, Republic of Korea,
\\
   \texttt{\{%
   \href{mailto:ssgyejin@yonsei.ac.kr}{ssgyejin},%
   \href{mailto:greghahn@yonsei.ac.kr}{greghahn},%
   \href{mailto:hsan@yonsei.ac.kr}{hsan},%
   \href{mailto:emmous@yonsei.ac.kr}{emmous}%
   \}@yonsei.ac.kr}
}

\begin{document}
\maketitle
\begin{abstract}
Implicit hate speech involves subtle and indirect expressions of prejudice or hostility toward a group.
Detecting it is challenging because it relies on nuanced context and implication rather than explicit offensive language.
Current approaches rely on contrastive learning, which is shown to be effective
on distinguishing hate and non-hate sentences.
Humans, however, detect implicit hate speech by first identifying specific
\emph{targets} within the text and subsequently interpreting how
these targets relate to their surrounding context.
Motivated by this reasoning process, we propose AmpleHate,
a novel approach designed to mirror human inference for implicit hate detection.
AmpleHate identifies explicit targets using a pre-trained Named Entity Recognition model
and captures implicit target information via [CLS] tokens.
It computes attention-based relationships between explicit, implicit targets and
sentence context and then,
directly injects these relational vectors into the final sentence representation.
This amplifies the critical signals of target-context relations for determining implicit hate.
Experiments demonstrate that AmpleHate achieves state-of-the-art performance, outperforming
contrastive learning baselines by an average of 82.14\% and
achieves faster convergence.
Qualitative analyses further reveal that attention patterns produced by AmpleHate
closely align with human judgement, underscoring its interpretability and robustness.
Our code is publicly available at: \url{https://github.com/leeyejin1231/AmpleHate}
\end{abstract}

\section{Introduction}\label{sec:intro}
\textbf{Warning}: \emph{this paper contains content that may be
offensive or upsetting}.\\

Internet and social media platforms have profoundly reshaped contemporary communication,
enabling rapid information dissemination and interpersonal interactions.
Yet, alongside the benefits, these platforms also enable the spread of hate speech,
becoming a serious social issue.
Recent studies indicate that approximately 30\% of young individuals have experienced cyberbullying~\citep{DusheJCNALSL23}.
This highlights the urgent need to detect and control hate speech.

For explicit hate speech detection,
traditional text classification approaches such that
learn global sentence representations or keyword features
are frequently used to distinguish hate from non-hate~\citep{GaoKH17, saleem2017web}.
These approaches have been successful as
explicit hate speech, with its direct and unambiguous characteristics,
contains clear expressions such as offensive words or insults.
Yet, implicit hate speech hides its hostility in latent context rather than explicit forms.
Specifically, it conveys hate through context, sarcasm, indirect expressions, or culturally specific references~\citep{WiegandRE21, ElSheriefZMASCY21}.
Consequently, standard keyword-based or holistic embedding methods often struggle to accurately detect implicit forms of hate.

\begin{figure}[t]
  \includegraphics[width=\columnwidth]{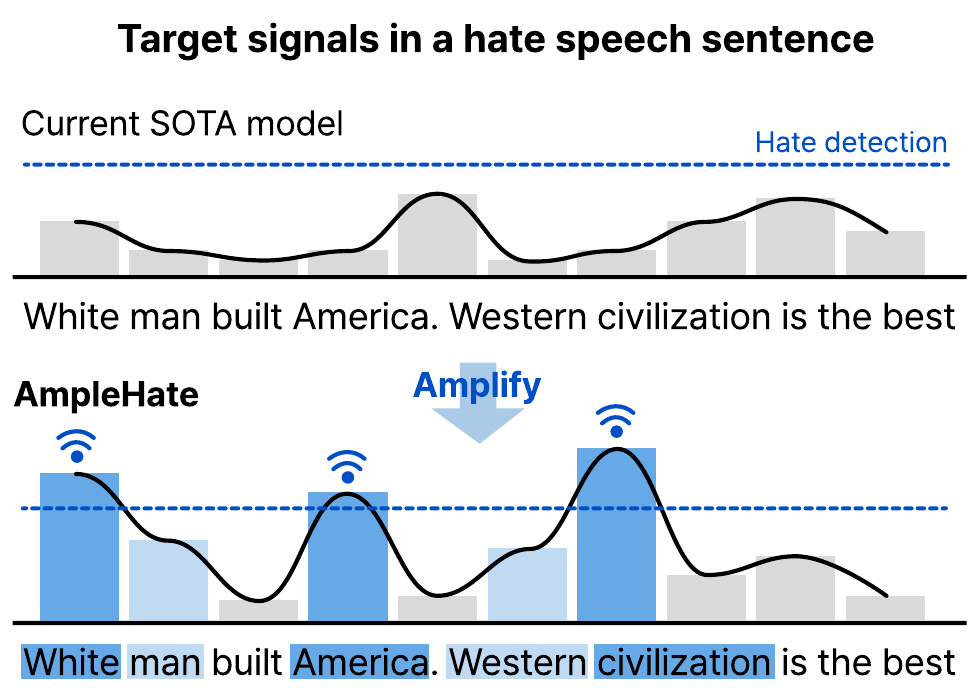}
  \caption{AmpleHate effectively detects implicit hate sentences by amplifying the target signals of hate-related tokens in the context of implicit hateness.}
  \label{fig:introduction}
\end{figure}

Recent studies have advanced implicit hate speech detection by employing contrastive learning frameworks,
such as InfoNCE-based objectives~\citep{oord2018representation, KimJPPH24}.
These methods lead semantically hateful sentences to have similar representations and push non-hateful sentences apart~\citep{KimPH22,AhnKKH24}.
Although these techniques are effective in implicit hate speech detection,
they operate on holistic sentence-level embeddings and overlooks the internal interactions within the sentences.

In contrast, humans identify implicit hate through a different and inherently more structured reasoning process:
first identifying specific \emph{targets}~(e.g.,
sensitive entities, which are demographic groups such as immigrants or Muslims, LGBTQ+ individuals, or cultural referents)
within a sentence and subsequently interpreting how the context frames or characterizes these
targets to infer hateful intent.
Motivated by this insight, we propose \textbf{AmpleHate},
a novel method explicitly designed to mirror the human inferential process of identifying implicit hate speech detection.

AmpleHate first identifies target entities~(explicit targets) within sentences using a pre-trained
Named Entity Recognition~(NER) model.
We then leverage the sentence-level embedding through the Transformer~\citep{VaswaniSPUJGKP17} encoder's [CLS] token
as an implicit representation of global contextual information~(implicit targets).
AmpleHate computes attention-based relational interactions between explicit, implicit targets and sentence contexts.
quantifying the influence each target has on the sentence's hateful meaning.
These relational vectors are then \emph{directly injected} into the final sentence embedding,
amplifying target-context signals and steering the classifier to base its decisions on
explicit target-context relationships, similar to human reasoning.

One might question the effectiveness of direct injection due to its susceptibility to potential noises.
However, by limiting the injection specifically to the relationships between explicit, implicit targets and context,
our method effectively enhances the learning of implicit hate speech.
As a result, AmpleHate achieves optimal performance earlier compared to existing contrastive learning approaches.

Through extensive experiments on implicit hate speech benchmarks,
AmpleHate achieves state-of-the-art performance, outperforming conventional contrastive learning baselines
by 6.87\%p in average.
Additionally, Figure~\ref{fig:introduction} and our qualitative analyses confirm that AmpleHate produces attention patterns closely aligned with
human judgement behaviors, highlighting its interpretaiblity and robustness.

\section{Related Work}

\subsection{Implicit Hate Speech}

Implicit hate speech targets protected groups
and is conveyed through subtle or indirect language rather than explicit 
slurs \citep{GNET2024}.
Unlike explicit hate, which can often be caught via keyword matching \citep{WarnerWHJ2012, DavidsonWMW17}
or simple toxicity filters \citep{WaseemH16,NobataTTMC16}, 
implicit hate relies on metaphor or context-dependent
cues that evade straightforward lexical methods.
This poses a significant challenge for automated implicit hate speech detection.
Recent works have primarily focused on two trajectories to address this challenge:
(1)~contrastive representation learning \citep{KimPNH23, AhnKKH24, KimJPPH24};
and (2)~data sampling/selection strategies \citep{OcampoCV23, KimAKH25, KimL2025}.
Meanwhile, these approaches depend on the availability of corpora specifically annotated for implicit hate speech,
such as \citet{SapGQJSC20, MathewSYBG021} and \citet{ElSheriefZMASCY21}.

However, most traditional studies treat the protected target merely as a class label
and thus fail to capture group-specific cues or coded language.
This oversight motivates the next subsection, which surveys target-aware 
modeling techniques that explicitly integrate target information.

\subsection{Target-Aware Modeling and Embeddings}
Recent methods not only identify the target~\citep{JafariAS24} but also
explicitly model target identity by using multi-task approach~\citep{ChirilPBMP22}.
They jointly learn to detect hatefulness and to predict the specific target category or 
identity group, leveraging shared representations and 
affective lexical to transfer knowledge across topics and improve generalization.
Adversarial training methods such as~\citet{XiaFT20} removed spurious group cues
to strip dialectal markers from embeddings and reduce false positives.
\citet{TongDXMGH24} used a hypernetwork conditioned on target embeddings to generate filters that eliminate biased features.
These approaches achieved significant gains in both accuracy and fairness.
In addition to those target-aware methods, we draw inspiration from recent advances
in embedding-space control and attention dynamics.
\citet{HanXL0SJAJ24} applied a low‐rank linear transform to embeddings to steer model behavior
with minimal overhead and
\citet{barberoFAXCMR2025} showed that attention sinks prevent over‐mixing in deep transformers.
Motivated by their findings, we inject a lightweight, target-aware attention module 
that both steers the model toward group-specific signals 
and dedicates an extra attention head to capture subtle, implicit target cues.

\section{AmpleHate: Amplify the Attention}\label{sec:amplehate}
\begin{figure*}
    \centering
        \includegraphics[width=\textwidth]{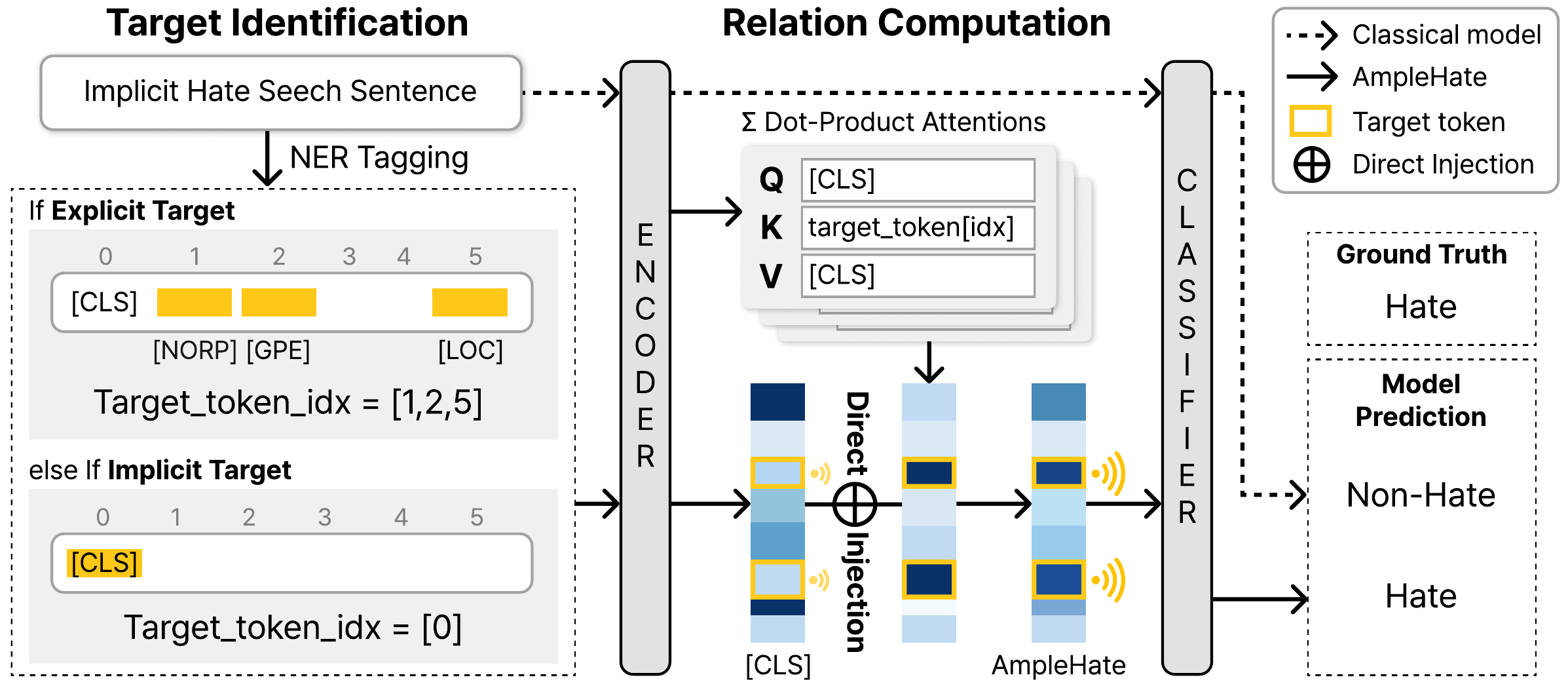}
        \caption{Overall procedure of AmpleHate.}
    \label{fig:overview}
\end{figure*}

We use the standard Transformer~\citep{VaswaniSPUJGKP17} self-attention mechanism in AmpleHate.
The multi-head attention calculation enables the model to represent sentence-level context more effectively.
AmpleHate consists of three core steps: 1) Target Identification, 2) Relation Computation,
and 3) Direction Injection.
Figure~\ref{fig:overview} illustrates the overall procedure of AmpleHate.

\subsection{Target Identification}\label{ssec:target_identification}
Let the input sentence~$X$ be tokenized as
\[
X = [x_0, x_1, \ldots, x_n], \quad x_0 = [\mathrm{CLS}],
\]
and let a Transformer encoder produce contextual embeddings
\[
H = [h_0, h_1, \ldots, h_n]\in \mathbb{R}^{(n+1)\times d},
\]
where $h_i \in \mathbb{R}^d$, $d$ is the size of hidden dimension,
and $h_0$ is the [CLS] embedding.
We begin by selecting the tokens that will serve as our \emph{explicit targets}.

In the implicit hate speech dataset, the critical entities are mainly groups, organizations and local~\citep{KhuranaNF25}.
Thus, we apply a pre-trained NER tagger to each token and retain only those with labels in
$\{\mathrm{ORG},\mathrm{NORP},\mathrm{GPE},\mathrm{LOC},\mathrm{EVENT}\}$,
yielding a binary mask $m_i\in {0,1}$ where $1\le i\le n$.
ORG represents organizations,
NORP represents nationalities, religious, and political groups,
GPE represents  countries, cities, and states,
LOC represents non-GPE locations such as mountain ranges or bodies of water,
and EVENT represents named events such as wars, sports events, or disasters.
Let
\[
I_{exp} = \{i\mid m_i = 1\}.
\]
The explicit target embeddings is then,
\[
H_{exp} = [h_i]_{i\in I_{exp}} \in \mathbb{R}^{|I_{exp}|\times d}.
\]
These embeddings~$H_{exp}$ serve as our explicit targets.
On the other hand, we use the embedding~$h_0$ of the [CLS] token
as our \emph{implicit token}~$H_{Imp}$.
This is supported by ~\citet{barberoFAXCMR2025} showing that initial tokens such as $[CLS]$ or $<bos>$
naturally attract significant attention and serve as stable anchors 
in transformer models, helping to regulate information.

\subsection{Relation Computation}\label{ssec:relation_compute}
Rather than simply boosting attention weights during training--which can fail to reflect the weights
directly to the final decision--AmpleHate aims to capture the fine-grained influence of
explicit targets~$H_{exp}$ and implicit targets~$H_{imp}$ on the context of the whole sentence~$X$.
We achieve this by applying a standard attention mechanism over the [CLS] embedding and the target embeddings,
producing a relational vector~$r\in \mathbb{R}^d$ that highlights the most relevant target-context interactions.

First, we form our attention inputs by letting the [CLS] embedding~$h_0$ act as both
the query and the value, and the target embeddings serve as keys:
\[
Q = h_0,\; K=H_{tgt}, \; V=h_0,
\]
where $H_{tgt}$ is either $H_{exp}$ or $H_{imp}$.
We then compute a score for each target $h_t\in H_{tgt}$ using scaled dot-product attention:
\[
r_{tgt}
= \sum_{k\in H_{tgt}} \mathrm{Softmax}(\frac{Qk^T}{\sqrt{d}})V.
\]
This dynamically assigns higher weights to targets whose embeddings align more closely with the sentence-level context.
As we compute when $H_{tgt} = H_{exp}$ and $H_{tgt} = H_{imp}$,
\[
r = r_{exp} + r_{imp}.
\]
Note that when we compute the attention for an implicit target,
the equation is the same as computing the self-attention of the [CLS] embedding.

The relational vector~$r$ thus encodes how each explicit and implicit target interacts with the
sentence as a whole by incorporating a learned self-adjustment of the global context of the sentence.

\subsection{Direct Injection}\label{ssec:direct_injection}
In order to explicitly amplify and reflect the target-context relation we have computed in Section~\ref{ssec:relation_compute},
we inject the relation vector~$r$ directly into the output, [CLS] embedding.
This \emph{direct injection} makes sure that the output contains the signals of target relations.
The updated output~$z$ is then passed to the classification head, ensuring that $z$ guides the right prediction~$\hat{y}$:
\[
z = h_0 + \lambda \cdot r, \quad \hat{y} = \mathrm{Softmax}(z),
\]
where $\lambda$ controls the degree of amplification for the injected signal.

This direct injection appropriately amplifies the signals of target-context relations for implicit hate speech detection,
yielding two key benefits:
\begin{description}
\item[Stronger decision ques.] By injecting $r$ into $h_0$ right before classification,
AmpleHate amplifies target-context interactions directly in the final logits.
This ensures that the AmpleHate's predictions rely on these amplified cues rather than only on
diffuse, indirect attention patterns.
\item[Selective noise reduction.] Since $r$ is built only from identified targets,
injection amplifies meaningful relations while filtering out irrelevant signals.
This effectively reduces potential noises and guarantees the faster convergence.
\end{description}

\subsection{Implementation Details}\label{ssec:implementation}
We identify the explicit targets by tagging named entities.
Specifically, we apply the pre-trained NER taggers~`dbmdz/bert-large-cased-finetuned-conll03-english' and `dbmdz/bert-base-cased-finetuned-conll03-english'.
Also, the degree~$\lambda$ of amplification in Section~\ref{ssec:direct_injection} is in $\{0.5, 0.75, 1.0, 1.25, 1.5\}$.

\section{Experiments and Analysis}

\begin{table*}
\centering
    \resizebox{\textwidth}{!}{
        \begin{tabular}{l|ccccccc|c}
        \noalign{\hrule height 0.8pt}
         & \multicolumn{7}{c|}{\small{\textbf{Datasets}}} & \multirow{2}{*}{Average} \\ 
        \makecell{\small{\textbf{Models}}} & IHC & SBIC & DYNA & Hateval & Toxigen & White & Ethos\\ 
        \noalign{\hrule height 0.8pt}
        BERT & 77.70 & 83.80 & 78.80 & 81.11 & 90.06 & 44.78 & 70.67 & 75.27\\
        SharedCon  & 78.50 & \textbf{84.30} & 79.10 & 80.24 & 91.21 & 46.15 & 69.05 & 75.50\\
        LAHN  & 78.40 & 83.98 & 79.64 & 80.42 & 90.42& 47.85 & 75.26 & 76.56\\
        \cdashline{1-9}
        AmpleHate (\emph{bert-base-ner}) & 81.46 & 83.79 & 81.39 & 80.56 & 90.74 & 71.96 & \textbf{78.61} & 81.21\\
        AmpleHate (\emph{bert-large-ner}) & \textbf{81.94} & 84.03 & \textbf{81.51} & \textbf{82.07} & \textbf{93.21} & \textbf{75.17} & 77.06 & \textbf{82.14}\\
        \noalign{\hrule height 0.8pt} 
        \end{tabular}
    }
\caption{Experimental results of baselines and ours for each dataset. 
\emph{bert-base-ner} means using a fine-tuned bert-base model, while \emph{bert-large-ner} means using a fine-tuned bert-large model for NER tagging.
The performance score~(macro-F1) is the average 
of three runs with different random seeds.}
\label{tab:in-domain}
\end{table*}

\begin{table*}
\centering
    \begin{tabular}{l|ccccccc|c}
    \noalign{\hrule height 0.8pt}
     & \multicolumn{7}{c|}{\small{\textbf{Test Datasets}}} & \multirow{2}{*}{Average} \\ 
    \makecell{\small{\textbf{Models}}} & IHC & SBIC & DYNA & Hateval & Toxigen & White & Ethos\\ 
    \noalign{\hrule height 0.8pt}
    BERT & 78.53 & 80.78 & \textbf{81.06} & 85.73 & 92.61 & 58.13 & 67.92 & 77.82\\
    SharedCon & 78.98 & 80.77 & 79.31 & 86.81 & 93.33 & 55.91 & 69.23 & 77.76\\
    LAHN & 77.40 & 80.34 & 79.64 & 84.49 & 90.48 & 59.09 & 68.34 & 77.11\\
    \cdashline{1-9}
    AmpleHate~(\emph{ours}) & \textbf{86.14} & \textbf{82.03} & 79.92 & \textbf{87.17} & \textbf{97.04} & \textbf{71.41} & \textbf{71.30}& \textbf{82.18}\\
    \noalign{\hrule height 0.8pt}
    \end{tabular}
\caption{Experimental results of baselines and ours for the whole dataset together as `combined' in Table~\ref{app-tab:data}.
We combine all datasets and treat them as a single dataset to see the general performance. We use \emph{bert-large-ner} model for NER tagging.
The performance score (macro-F1) is the average 
of three runs with different random seeds for the combined dataset.}
\label{tab:OOD1}
\end{table*}

\subsection{Datasets}

\paragraph{IHC}\noindent\citep{ElSheriefZMASCY21} is designed to support implicit hate speech classification, 
featuring about 22k tweets annotated not only with hate labels 
but also with textual descriptions that reveal underlying hateful implications.
\paragraph{SBIC}\noindent\citep{SapGQJSC20} offers a large-scale resource for studying social bias in language, 
pairing social media posts with structured frames that annotate offensiveness, speaker intent, and group targets.
\paragraph{Dynahate}\noindent\citep{VidgenTWK20} introduced a challenging hate speech dataset built 
via a human-and-model-in-the-loop process, incorporating perturbed and implicitly hateful examples 
to enhance model generalization and robustness.
\paragraph{Hateval}\noindent\citep{BasileBFNPPRS19} contains about 19k Twitter posts annotated for 
binary hate speech specifically targeting immigrants or women, 
with additional tags for aggressiveness and target scope. 
\paragraph{Toxigen}\noindent\citep{HartvigsenGPSRK22} comprises approximately 274k machine-generated statements, 
evenly split between toxic and non-toxic language concerning 13 minority groups.
\paragraph{White}\noindent\citep{GibertPGC18} is a hate speech dataset collected from a White Supremacy Forum. 
Each sentence is manually annotated for the presence or absence of hate speech 
according to specific annotation guidelines.
\paragraph{ETHOS}\noindent\citep{MollasCKT22} offers 998 YouTube/Reddit comments labeled 
for hate speech presence or absence. 
Table~\ref{app-tab:data} in Appendix~\ref{app:data_statistics}
reports the data statistics.

\subsection{Baselines}
\paragraph{BERT}~\citep{DevlinCLT19} is a transformer-based language model pre-trained on large corpora
using masked language modeling and next sentence prediction. Its bidirectional encoding enables effective
contextual understanding for downstream classification tasks.
\paragraph{SharedCon}~\citep{AhnKKH24} is a SOTA method for the implicit hate speech detection task. 
The approach clusters sentence embeddings by label, uses each cluster centroid as an anchor, 
and then fine-tunes the model with a supervised contrastive loss.
\paragraph{LAHN}~\citep{KimJPPH24} integrates momentum contrastive learning 
with label-aware hard-negative sampling, selecting the most similar opposite-label sampling 
for each anchor and training with a joint supervised-contrastive and cross-entropy loss.

\subsection{Experimental Setup}
We use the BERT-base model~\citep{DevlinCLT19} with the 100M parameter size as our encoder and train on an NVIDIA RTX 4090 GPU.
The decision threshold ranges from 0.05 to 0.95 in steps of 0.05.
We evaluate model performance using the macro-F1 and report
averaged scores over three independent runs with different random seeds.
We use the AdamW optimizer with a fixed learning rate of 2e-5, a batch size of 16, and 6 training epochs.

\subsection{Experimental Results}\label{ssec:baseline_comparison}
We begin with experiments on individual datasets, where each model is trained and evaluated on the same dataset.
We then train a single model on the combined dataset and evaluate it across individual test sets, 
demonstrating the versatility of our method.

\paragraph{Individual Dataset.}
Table~\ref{tab:in-domain} shows that
AmpleHate consistently outperforms nearly all baseline models, 
with particularly strong gains on datasets with implicit hate (e.g., +27.32\%p on White). 
On average, it improves over BERT by 6.87\%p and over the strongest baseline (LAHN) by 5.58\%p.

These results confirm that identifying explicit and implicit targets,
coupled with amplification of target-context relation signals by direct injection,
yields clear and consistent improvements over contrastive learning approaches.

\paragraph{Combined Dataset.}
\label{sec:combined_dataset}

In this setup, the model is trained on the combined set of all seven datasets and then
evaluated independently on each test set.

Table~\ref{tab:OOD1} shows that AmpleHate achieves the highest macro-F1 on six out of seven datasets,
gaining approximately 5\%p gains on average for all baselines.

These findings indicate that AmpleHate maintains its effectiveness even without dataset-specific fine-tuning,
making it suitable for practical deployment.
By incorporating target-focused attention during training, the model retains robustness across
varying expressions of implicit hate speech.

\subsection{Case Analysis}

\begin{figure}[t]
  \includegraphics[width=\columnwidth]{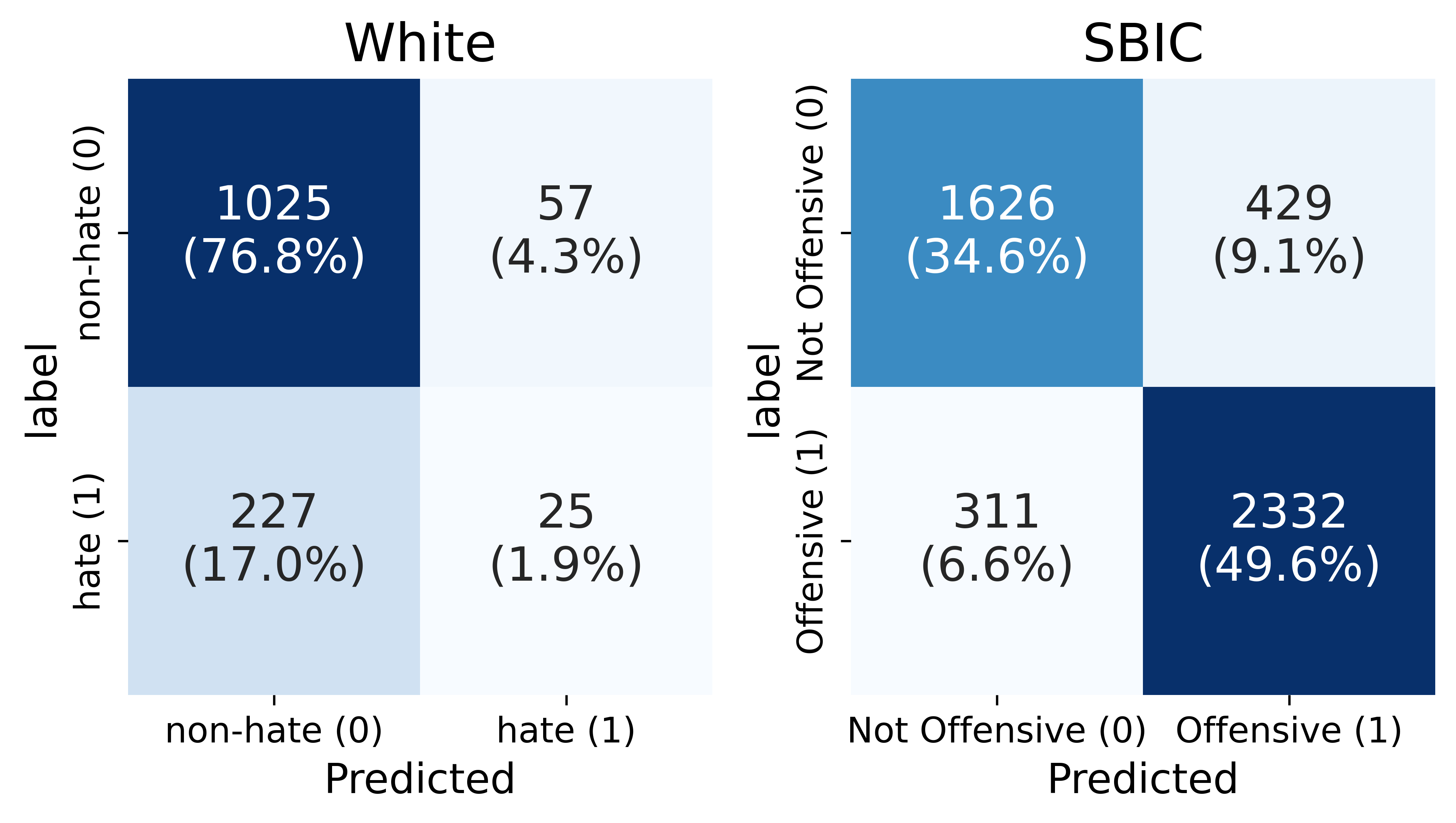}
  \caption{Confusion matrices for the White (left) and SBIC (right) datasets, 
  showing counts and percentages of true vs. predicted labels for `non-hate/not-offensive' and `hate/offensive' classes.}
  \label{fig:confusion}
\end{figure}

\paragraph{Analysis on White Dataset Characteristics}

In the individual dataset experiments, AmpleHate achieves the highest relative improvement on the White dataset, which motivates a detailed examination of its linguistic and structural characteristics.
Figure~\ref{fig:confusion} shows the confusion matrix on White (left),
where AmpleHate demonstrates strong discrimination between hate and non-hate instances, despite the skewed label distribution.
The model maintains high precision on non-hate samples while successfully identifying a portion of subtle hate cases.

Several factors likely contribute to this result.
First, White contains a high proportion of implicit hate speech,
where hateful intent emerges through context rather than direct slurs.
Many examples are structured in a way that requires identifying the target entity
before the hatefulness becomes clear.
This aligns well with the intended role of AmpleHate--to highlight target mentions and capture context-dependent signals.

Second, explicit references to target groups frequently appear in the data.
While standard models may treat these as ordinary nouns,
AmpleHate interprets them as meaningful cues and allocates attention accordingly.

These properties make the White dataset particularly well-matched to the goal behind
AmpleHate, which focuses on detecting subtle and indirect hate speech through target-aware contextual understanding.

\paragraph{Error Analysis on SBIC}

In Table~\ref{tab:in-domain}, AmpleHate shows only a marginal improvement
over the baseline on the SBIC dataset.
We analyze the SBIC dataset to understand this experimental result.

Hate speech datasets usually involve explicit attacks against specific identity groups.
In contrast, offensive language datasets label content as offensive regardless of whether the language targets a group.
SBIC adopts this broader definition, which creates a fundamental distinction from 
the other datasets in the study.

Figure~\ref{fig:confusion} shows the confusion matrix for SBIC (right).
We extract 740 misclassified examples and analyze their content.
Most errors fall into the false positive category.
Many of these cases contain highly offensive expressions,
but the original label marks them as non-offensive.
For example, the sentence "in fucking mexico bitches" includes profanity and
a strong negative tone, but it carries a non-offensive label.

We randomly sample 100 of the misclassified examples and review them manually.
Among them, 51\% appear to contain labeling inconsistencies.
This high noise ratio suggests that SBIC includes annotation artifacts that
differ from hate speech corpora.
These inconsistencies likely contribute to the limited performance gains of AmpleHate on SBIC.

\section{Analysis}
\label{sec:analysis}

\begin{table*}
\centering
\resizebox{\textwidth}{!}{
\begin{tabular}{l|c|c|c|c|c|c|c|c}
\noalign{\hrule height 0.8pt}
 & \multicolumn{7}{c|}{\textbf{Datasets}} & \\
\small{\textbf{Target}} & IHC & SBIC & DYNA & Hateval & Toxigen & White & Ethos & Avg. (F1) \\
 & F1 (step) & F1 (step) & F1 (step) & F1 (step) & F1 (step) & F1 (step) & F1 (step) & \\
\noalign{\hrule height 0.8pt}
No Target & 77.70 \small{(1146)} & 83.80 \small{(2016)} & 78.80 \small{(2316)} & 81.11 \small{(1953)} & 90.06 \small{(1147)} & 44.78 \small{(\textbf{360})} & 70.67 \small{(160)} & 75.27\\
Random Target & 77.00 \small{(\textbf{450})} & 83.95 \small{(\textbf{750})} & 79.99 \small{(1950)} & 79.44 \small{(400)} & 89.59 \small{(\textbf{300})} & 59.46 \small{(500)} & 67.54 \small{(\textbf{50})} & 76.71 \\
Implicit Target & 78.76 \small{(1200)} & 83.63 \small{(2167)} & 80.54 \small{(\textbf{647})} & 81.70 \small{(2120)} & 90.52 \small{(647)} & 59.00 \small{(413)} & 75.57 \small{(160)} & 78.53\\
Explicit Target & 77.92 \small{(910)} & 83.93 \small{(2150)} & 80.98 \small{(907)} & 81.33 \small{(1827)} & 90.55 \small{(907)} & 59.69 \small{(387)} & 72.33 \small{(120)} & 78.10\\
Implicit + Explicit & \textbf{81.93} \small{(1246)} & \textbf{84.03} \small{(2166)} & \textbf{81.51} \small{(2350)} & \textbf{82.07} \small{(\textbf{1647})} & \textbf{93.21} \small{(933)} & \textbf{75.16} \small{(380)} & \textbf{77.06} \small{(127)} & \textbf{82.13}\\
\noalign{\hrule height 0.8pt}
\end{tabular}
}
\caption{Ablation study results showing macro-F1 scores and stopping steps for five target configurations (No Target, Random Target, Implicit Target, Explicit Target, Implicit + Explicit) across seven datasets. 
We use \textit{bert-large-ner} tagger to explicit tagging.
Stopping epochs are averaged over three runs and then rounded to the nearest integer.}
\label{tab:ablation_extended}
\end{table*}

\begin{table*}[t]
\centering
\renewcommand{\arraystretch}{1.3} 
\begin{tabularx}{\textwidth}{lX}
\noalign{\hrule height 0.9pt}
\textbf{Dataset} & \textbf{Highlighted Implicit Hate Speech Sentence} \\
\hline
IHC     & even the \colorbox{lightblue}{white nationalist} movement is ran by jews. \\
\hline
SBIC    & Yo \colorbox{lightblue}{hoe} is like a tick so she \colorbox{lightblue}{hunting} for a \colorbox{lightblue}{blood} \\
\hline
DYNA    & tell \colorbox{lightblue}{chicks} to stop \colorbox{lightblue}{dressing} like street walkers \\
\hline
Hateval & Arrest \colorbox{lightblue}{them} and \colorbox{lightblue}{DEPORT} THEM ALL!!!! \#\colorbox{lightblue}{BuildThatWall} \\
\hline
Toxigen & all \colorbox{lightblue}{black} people are suspected of trying to pull something \\
\hline
White   & The \colorbox{lightblue}{German men} sound so sexy. \\
\hline
Ethos   & Too bad \colorbox{lightblue}{women} don’t know how to kill \colorbox{lightblue}{themselves} \\
\noalign{\hrule height 1pt}
\end{tabularx}
\caption{Visualization of token‐level attention weights derived from the [CLS] token for each dataset example.
Highlighted tokens are those that the AmpleHate model considers more important than a standard BERT baseline, 
in the implicit‐target scenario where no explicit target token is provided. 
}
\label{tab:sentence}
\end{table*}

\subsection{Impact of Targets}\label{ssec:target_impact}

AmpleHate distinguishes between random targets, explicit targets, identified via NER tagging,
and implicit targets represented by the [CLS] token, which captures global context.
We conduct an ablation study to understand the contribution of each target type.
The experiment includes fine-tuning the vanilla BERT baseline without any target signals,
with randomly assigned 20\% targets, with only the implicit targets, with only explicit targets, and with both implicit and explicit targets combined.
Since explicit targets were tagged for approximately 20\% of the data on average, we introduced the Random 20\% condition to serve as a comparable baseline.
Table~\ref{tab:ablation_extended} presents both the macro F1-scores and the number of convergence steps
across these settings.

Interestingly, our results reveal that incorporating only the implicit target yields
greater improvements than relying solely on explicit targets.
In particular, using the implicit targets shows 3.26\%p improvements while
using the explicit targets shows 2.83\%p on average, compared to fine-tuning BERT with no signals.
This finding suggests that, in the case of implicit hate speech--where the intent
is often in latent form and not tied to named entities--contextual
representation via the [CLS] token is particularly powerful for target-context relation computation.
Nevertheless, models leveraging explicit targets also outperform the baseline without any target signals,
confirming that entity-specific cues add valuable information for detecting implicit hate.
Crucially, when both implicit and explicit targets are integrated,
AmpleHate achieves the highest performance across all settings.

In addition to the macro-F1 performance, we observe the training efficiency.
Including target signals generally achieves faster convergence steps than fine-tuning vanilla BERT with no signals.
On average, using only implicit or only explicit targets reduces the number of steps for the convergence by approximately 20\%.
However, when both target types are used together,
the reduction in convergence steps is more modest with approximately 3\% compared to the vanilla BERT.
This small tradeoff is caused by a substantial gain in detection performance, achieving 6.86\%p improvement.

These results suggest that while incorporating implicit or explicit targets individually
leads to faster convergence but slight performance gains,
combining both enables AmpleHate to achieve the best performance along with the competitive convergence speed.

\subsection{Comparison of Convergence Step}

\begin{figure}[t]
  \includegraphics[width=\columnwidth]{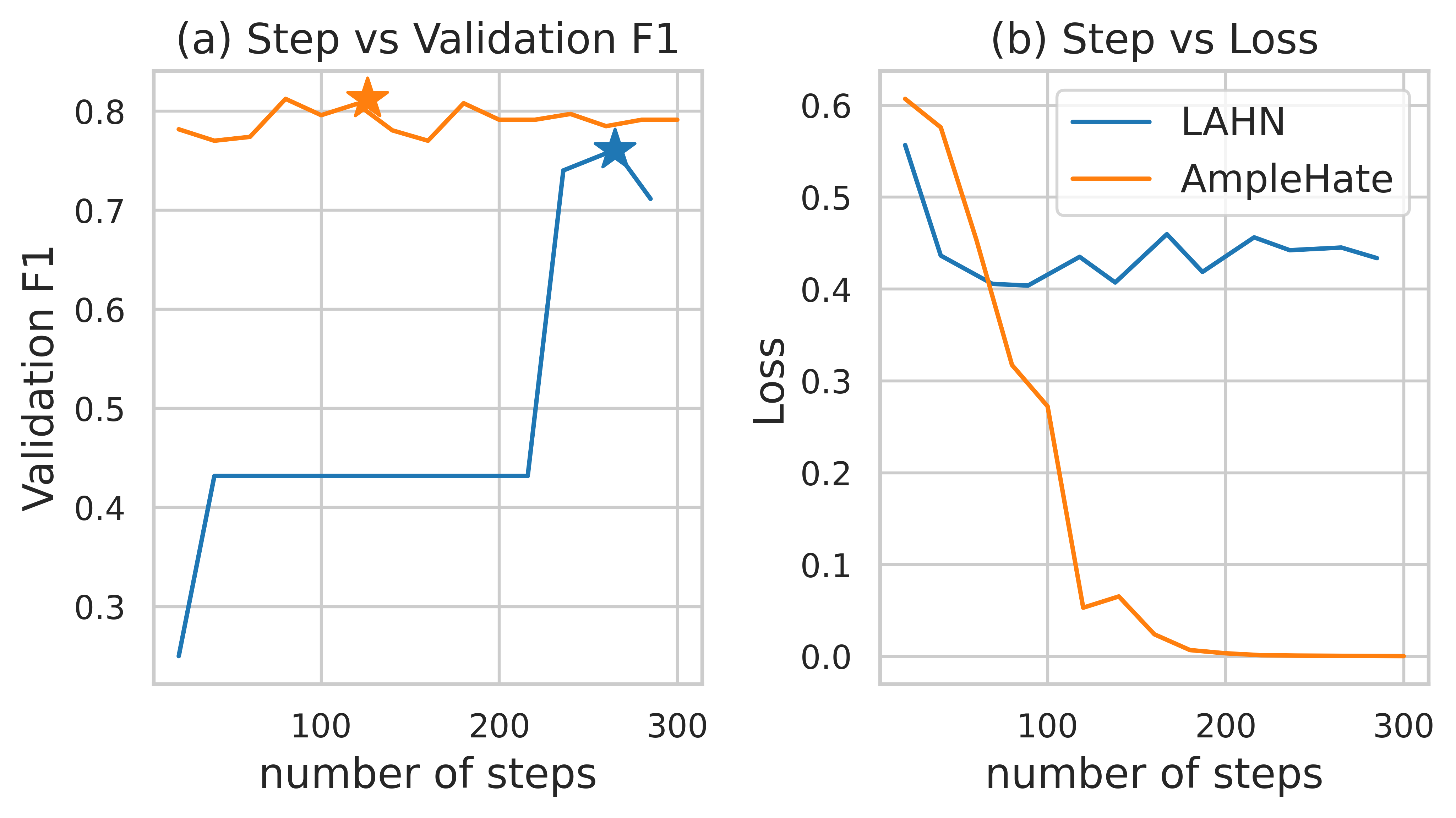}
  \caption{Comparison of AmpleHate and LAHN on the Ethos datasets.
  (a) Validation F1-score over training steps, with start indicating the stopping step.
  (b) Training loss over steps. AmpleHate converges faster and maintains lower loss than LAHN.
  Appendix~\ref{app:convergence_step} provides further details.
  }
  \label{fig:step}
\end{figure}

We compare the convergence steps of AmpleHate with the state-of-the-art model LAHN
to assess the training efficiency and optimization stability of our approach.
Figure~\ref{fig:step} presents the comparison on the Ethos dataset.
Figure~\ref{fig:step}~(a) shows the validation F1-score each training step,
where the star markers indicate the step at which each model achieves its highest validation performance.
AmpleHate reaches its peak F1-score at an earlier step than LAHN,
achieving comparable or better performance with significantly fewer training steps.
Figure~\ref{fig:step}~(b) represents the loss curves over the same training steps,
where AmpleHate consistently maintains a lower loss throughout training.
It demonstrates that AmpleHate is more stable and effective for optimization.

This early convergence and stable training behavior indicate that AmpleHate is more computationally
efficient and less prone to overfitting compared to LAHN.
Similar patterns appear across other datasets, with detailed results provided in Appendix~\ref{app:convergence_step}.
These findings highlight the strength of AmpleHate in delivering high performance
with rapid and stable convergence, making it a highly efficient solution for implicit hate speech detection.

\subsection{Token-level Attention with Implicit Tokens}\label{ssec:token_level_attention}

\begin{figure}
    \centering
        \includegraphics[width=\columnwidth]{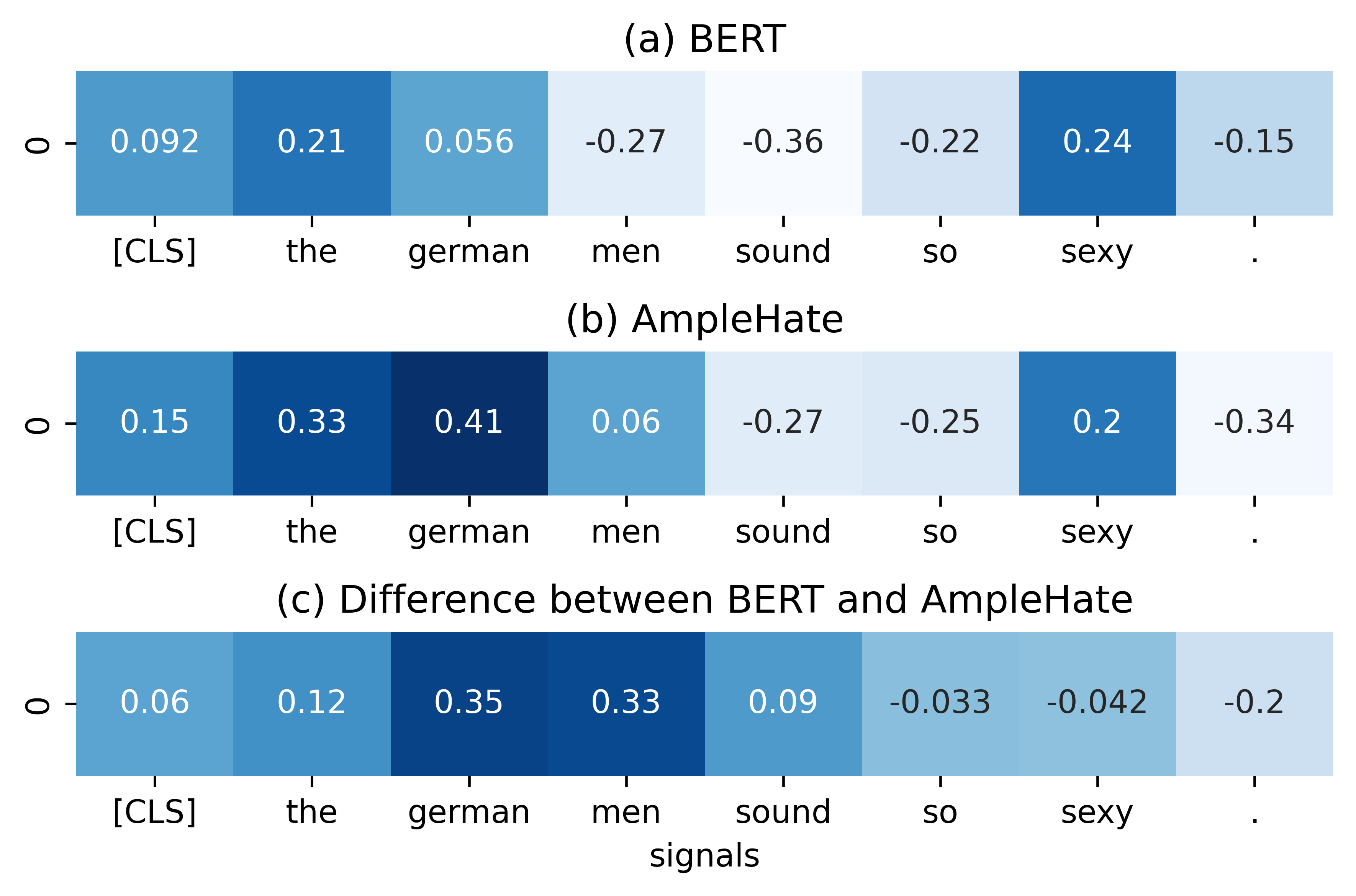}
        \caption{Token-level target signals between the [CLS] representation and each token in an implicit hate speech sentence from the White dataset. (a) shows BERT's signals, (b) shows signals from AmpleHate trained on combined datasets and (c) highlights the token-level difference.}
    \label{fig:white_output}
\end{figure}

AmpleHate assigns the [CLS] token as a target token
when no explicit target entity appears in a sentence.
This strategy raises a question: does [CLS] effectively serve as an
implicit focus point in such cases?
To answer this question, we examine how strongly the [CLS] representation
attends to each token in sentences labeled as implicit hate speech.

Figure~\ref{fig:white_output} visualizes token-level target signals with [CLS] for
three cases: (a) BERT, (b) AmpleHate, (c) the difference between them.
We use AmpleHate trained on the combined\_datasets setting described in Section~\ref{sec:combined_dataset}.
BERT represents attention broadly, without clear focus.
In contrast, AmpleHate places greater emphasis on the token "germen".
This token plays a central role in interpreting the hateful intent of the sentence.
Even without target supervision, AmpleHate identifies "germen" as a core indicator of implicit hate, suggesting that its target-aware attention mechanism successfully captures subtle targets.
This observation implies that computing token-level relation with [CLS] provides sufficient
signal for detecting implicit hate, even in the absence of explicit target tokens.

Table~\ref{tab:sentence} extends the analysis across all seven datasets.
In each example, we highlight tokens where AmpleHate assigns significantly higher attention than BERT. 
These implicit targets often include stereotype triggers.
The results demonstrate that AmpleHate consistently prioritizes tokens humans are
also likely to attend to when interpreting implicit hate.

This analysis confirms that using [CLS] as the target token does not disrupt the model's ability.
Instead, it enables AmpleHate to maintain robust attention over hate-relevant context, even without explicit target supervision.

\section{Conclusion}\label{sec:conclusion}
We introduced AmpleHate, an approach that mirrors human reasoning for
implicit hate detection by focusing on target-context interaction,
in contrast to conventional contrastive learning approaches.
AmpleHate first
identifies explicit and implicit targets,
then computes attention-based relation vectors, and
finally injects the signals into the output embeddings.
This targeted injection amplifies target-context relations,
which specifically contain relevant signals for implicit hate, suppressing noise.
AmpleHate achieves the state-of-the-art F1 scores on seven
implicit hate speech datasets both
in individual and combined dataset train settings.
This empirically shows the generalizability of AmpleHate.
Furthermore, AmpleHate presents faster convergence than the
contrastive-learning baselines.
While AmpleHate depends on an external NER tagger to locate explicit targets,
we aim to explore richer target detection and relation injection strategies
as well as applications to other tasks that hinge on subtle target-context reasoning.
Overall, AmpleHate demonstrates that encoding and injecting target-context relations
can outperform conventional contrastive learning methods,
offering a more robust and efficient approach for implicit hate speech detection.

\section*{Limitations}

AmpleHate relies on the accuracy of the external NER tagger for explicit target identification;
errors or insufficient entity recognition may limit performance,
especially for context-specific or newly emerging hate expressions with latent implications.
In addition, the approach of AmpleHate on modeling implicit target centers on the [CLS] token,
which may not always capture the full range of implicit cues,
particularly in longer or more complex sentences.

Moreover, while AmpleHate demonstrates strong performance in both macro-F1 scores and the convergence rate,
there remains the possibility that AmpleHate may not perform the best on forms of implicit hate which are not well
represented in current benchmarks or in new domains.

We plan to address these limitations in future work by developing more adaptive and context-aware target strategies
and extending our approach to tasks that involve more complex or subtle forms of latent hate. 

\section*{Ethical Consideration}
\paragraph{Minimizing Annotator Harm}
Conventional hate speech detection approaches rely on manual annotations,
which can expose annotators to distressing or harmful sentences.
AmpleHate reduces this exposure by leveraging implicit contextual signals,
thus lowering the dependence on explicit and manual labels.
This contributes to a more ethical data collection and annotation pipeline
by minimizing the mental burden on human annotators~\citep{VidgenHNTHM19}.

\paragraph{Contextual Awareness}
AmpleHate models explicit and implicit target information,
enabling it to recognize diverse and nuanced forms of hate speech.
By capturing latent patterns through its design, AmpleHate supports
more context-sensitivity and fair detection,
reducing the risk of overfitting to majority groups or missing contextually-dependent harms.

\paragraph{Risks of Potential Misuse}
AmpleHate could be misused--such as by
adversarial users attempting to bypass detection or
abuse the method to generate more sophisticated hate speech.
The deployment of AmpleHate should be accompanied
by careful monitoring and critical assessment of model outputs to address these risks.

\section*{Acknowledgments}
This research was supported by the 
NRF grant~(RS-2025-0222262) and the AI Graduate School Program (RS-2020-II201361) funded by the Korean government~(MSIT).

\bibliography{main}
\newpage

\appendix

\section{Data Statistic}\label{app:data_statistics}

We evaluate AmpleHate on seven publicly available hate speech datasets covering both explicit and implicit targets.
Table~\ref{app-tab:data} summarizes the number of examples in each train, validation, and test split.
Note that the combined corpus’s test set (marked with $\dagger$) is held out from our main evaluations.

\begin{table}[h!]
    \centering
    \begin{tabular}{cccc}
    \noalign{\hrule height 0.8pt}
         \textbf{Dataset} & \textbf{Train set} & \textbf{Valid set} & \textbf{Test set} \\
    \noalign{\hrule height 0.4pt}
        IHC & 14,932 & 1,867 & 1,867 \\
        SBIC & 35,504 & 4,673 & 4,698 \\
        DYNA & 33,004 & 4,125 & 4,126 \\
        Hateval & 10,384 & 1,298 & 1,298 \\
        Toxigen & 5,420 & 678 & 678 \\
        White & 10,668 & 1,334 & 1,334 \\
        Ethos & 798 & 100 & 100 \\\midrule
        Combined & 110,710 & 14,075 & 14,101$^\dagger$ \\
    \noalign{\hrule height 0.8pt}
    \end{tabular}
    
    \caption{The statistical information of datasets in our experiments.
    The test split of the combined dataset (denoted with $\dagger$) is excluded from AmpleHate evaluations.
    }
    \label{app-tab:data}
\end{table}

\section{Convergence Efficiency}\label{app:convergence_step}
We compare the convergence behavior of AmpleHate and LAHN,
a state-of-the-art contrastive learning baseline across all datasets, presented in Table~\ref{app-fig:step}.
Here, convergence step refers to the validation step at which early stopping selects the best model.
Lower convergence steps indicate faster model training and more efficient progression toward optimal performance.

Acrross six out of seven benchmarks,
AmpleHate consistently reaches peak performance
with substantially lower training steps than LAHN.
For instance, AmpleHate converges more than three times faster on the White dataset,
and achieves approximately~$\times$ 1.5 speedup for the other datasets.
Notably, AmpleHate not only converges faster but also achieves higher macro-F1 scores
than LAHN reported in Table~\ref{tab:in-domain},
demonstrating that its efficiency does not degrade its performance.
The only exception is Hateval, where LAHN converges marginally faster.
However, as shown in Table~\ref{tab:in-domain}
AmpleHate outperforms LAHN by approximately 2\%p in macro-F1,
confirming that this tradeoff is reasonable.

\begin{table*}[h!]
\centering
    \begin{tabular}{c L ccccccc}
    \noalign{\hrule height 0.8pt}
     & \multicolumn{7}{c}{\small{\textbf{Datasets}}}\\ 
    \makecell{\small{\textbf{Models}}} & IHC & SBIC & DYNA & Hateval & Toxigen & White & Ethos\\ 
    \noalign{\hrule height 0.8pt}
    LAHN      & 1683 & 3969 & 2962 & 1358 & 1592 & 1286 & 256\\
    AmpleHate~\emph{ours} & 980  & 2250 & 2100 & 1646 & 933  & 380  & 126\\\specialrule{0.85pt}{0pt}{0pt}
    \rule{0pt}{2.5ex}
    $\mathrm{Speedup} = \frac{\mathrm{LAHN \; steps}}{\mathrm{AmpleHate\; steps}}$
              & $\times 1.72$ & $\times 1.76$ & $\times 1.41$ & $\times 0.83$& $\times 1.71 $ & $\times 3.38$ & $\times 2.03$\\
    \noalign{\hrule height 0.8pt}
    \end{tabular} 
\caption{Convergence results of AmpleHate and LAHN.
For each dataset, we report the early stopping step at which the best validation performance is achieved,
as well as the relative speedup~($\times$) of AmpleHate over LAHN.
A higher speedup indicates that AmpleHate converges in fewer steps, improving the training efficiency.
}
\label{app-fig:step}
\end{table*}

\begin{table*}[hbt]
\centering
\begin{tabular}{l L ccc L c R c}
\specialrule{0.85pt}{0pt}{0pt}
Dataset   & Run 1 & Run 2 & Run 3 & AVG. $\pm$ STD. & CI \\
\hline
IHC       & 82.31 & 81.63 & 81.87 & 81.94 $\pm$ 0.34 & [81.10, 82.78] \\
SBIC      & 84.05 & 84.15 & 83.90 & 84.03 $\pm$ 0.13 & [83.71, 84.35] \\
DYNA      & 81.44 & 81.77 & 81.32 & 81.51 $\pm$ 0.23 & [80.94, 82.08] \\
Hateval   & 82.14 & 82.02 & 82.04 & 82.07 $\pm$ 0.06 & [81.92, 82.22] \\
Toxigen   & 93.83 & 92.60 & 93.21 & 93.21 $\pm$ 0.62 & [91.67, 94.75] \\
White     & 74.62 & 75.88 & 75.00 & 75.17 $\pm$ 0.65 & [73.56, 76.78] \\
Ethos     & 79.14 & 77.29 & 74.76 & 77.06 $\pm$ 2.20 & [71.59, 82.53] \\\hline
AVG.      & 82.50 & 82.19 & 81.73 & 82.14 $\pm$ 0.39 & [81.17, 83.11] \\
\specialrule{0.85pt}{0pt}{0pt}
\end{tabular}
\caption{Experimental results of AmpleHate for each dataset.
Each row reports the results for a given dataset, including all three runs, the average~(AVG.), standard deviation~(STD.), and
the 95\% confidence interval~(CI) using the t-distribution.}
\label{tab:statistical-significance}
\end{table*}

The primary factor of AmpleHate's rapid convergence is its direct injection of
target-context relational signals.
By amplifying the most relevant cues of implicit hate by using targets,
AmpleHate enables the model to focus on the critical features earlier in training,
reducing unnecessary parameter updated and accelerating optimization.

Overall, these findings highlight that AmpleHate is both accurate
as well as highly efficient, requiring fewer training steps and converging more rapidly
than contrastive learning methods.
This strengthens AmpleHate's suitability for implicit hate speech detection.

\section{Statistical Significance Analysis}\label{sec:statstical_significance}
We evaluate the performance of AmpleHate across three random seeds on eight implicit hate speech detection datasets.
Table~\ref{tab:statistical-significance} reports the results including
the average and standard deviation for each dataset.
We compute a 95\% confidence interval~(CI) using the $t$-distribution
which is widely used to analyze statistical significance~\citep{KeithO18,Bestgen22,GladkoffHN23,BaillargeonL24,LiH00LSY24}.
Since we conducted 3 runs in total, our degrees of freedom~(df) is 2 and
the interval is calculated as follows:

\[
\mathrm{CI} = \bar{x} \pm t_{(\frac{\alpha}{2}, df)} \cdot \frac{s}{\sqrt{n}} = \bar{x} \pm t_{(0.025, 2)} \cdot \frac{s}{\sqrt{n}},
\]
where $\bar{x}$ is the average score, $s$ is the standard deviation,
$n$ is the number of runs, and $t_{0.025,2}\approx 4.303$ is the t-statistics for 95\% confidence
with two degrees of freedom.

The results demonstrate that AmpleHate achieves highly consistent performance across multiple runs.
For instance, the CI for SBIC and IHC is [83.71, 84.35] and [81.10, 82.78], respectively.
These findings underscore the reliability and robustness of AmpleHate.

\end{document}